\documentclass[acmsmall,nonacm=true,authorversion=true]{acmart}

\usepackage{comment}

\AtBeginDocument{%
  }

\usepackage{todonotes}
\usepackage{tabularx}
\usepackage{booktabs}
\usepackage{tikz}
\usepackage{forest}
\usetikzlibrary{trees,positioning,shapes,shadows,arrows.meta}
\usepackage{threeparttable}

\setcopyright{none}
\renewcommand\footnotetextcopyrightpermission[1]{}
\begin{document}
\makeatletter
\let\authorsaddresses\@empty      
\let\@authorsaddresses\@empty     
\let\@printauthorsaddresses\relax 
\makeatother
\title{Automatic Speech Recognition (ASR) for African Low-Resource Languages: A Systematic Literature Review}



\author{Sukairaj Hafiz Imam} 
\email{shafizimam@nwu.edu.ng}
\orcid{xxxx-xxxx-xxxx }
\affiliation{
    \institution{Northwest University Kano}
    \country{Nigeria}}
    
\author{Tadesse Destaw Belay}
\email{tadesseit@gmail.com}
\affiliation{
    \institution{Instituto Politécnico Nacional}
    \country{Mexico}}

\author{Kedir Yassin Husse} 
\email{kediryassin25@yahoo.com}
\orcid{xxxx-xxxx-xxxx }
\affiliation{
    \institution{University of Gondar}
    \country{Ethiopia}}
    
\author{Ibrahim Said Ahmad}
\email{isahmad.it@buk.edu.ng}
\affiliation{
    \institution{Bayero University Kano}
    \country{Nigeria}}
    
\author{Idris Abdulmumin}
\email{abumafrim@gmail.com}
\affiliation{
    \institution{University of Pretoria}
    \country{South Africa}}

\author{Hadiza Ali Umar}
\email{haumar.cs@buk.edu.ng}
\affiliation{
    \institution{Bayero University Kano}
    \country{Nigeria}}
    
\author{Muhammad Yahuza Bello}
\email{mybello.maths@buk.edu.ng}
\affiliation{
    \institution{Bayero University Kano}
    \country{Nigeria}}

\author{Joyce Nakatumba-Nabende}
\email{joyce.nabende@mak.ac.ug}
\affiliation{
    \institution{ Makerere University}
    \country{Uganda}}
    
\author{Seid Muhie Yimam}
\email{seid.muhie.yimam@uni-hamburg.de}
\affiliation{
    \institution{University of Hamburg}
    \country{Germany}}
    
\author{Shamsuddeen Hassan Muhammad}
\email{shamsuddeen2004@gmail.com}
\affiliation{
    \institution{Imperial College London}
    \country{United Kingdom}}



\begin{abstract}
Automatic Speech Recognition (ASR) has achieved remarkable global progress, yet African low-resource languages remain rigorously underrepresented, producing barriers to digital inclusion across the continent with more than +2000 languages. This systematic literature review (SLR) explores research on ASR for African languages with a focus on datasets, models and training methods, evaluation techniques, challenges, and recommends future directions. We employ the PRISMA 2020 procedures and search DBLP, ACM Digital Library, Google Scholar, Semantic Scholar, and arXiv for studies published between January 2020 and July 2025. We include studies related to ASR datasets, models or metrics for African languages, while excluding non-African, duplicates, and low-quality studies (score <3/5). We screen 71 out of 2,062 records and we record a total of 74 datasets across 111 languages, encompassing approximately 11,206 hours of speech. Fewer than 15\% of research provided reproducible materials, and dataset licensing is not clear. self-supervised and transfer learning techniques are promising, but are hindered by limited pretraining data, inadequate coverage of dialects, and the availability of resources. Most of the researchers use Word Error Rate (WER), with very minimal use of linguistically informed scores such as Character Error Rate (CER) or Diacritic Error Rate (DER), and thus with limited application in tonal and morphologically rich languages. The existing evidence on ASR systems is inconsistent, hindered by issues like dataset availability, poor annotations, licensing uncertainties, and limited benchmarking. Nevertheless, the rise of community-driven initiatives and methodological advancements indicates a pathway for improvement. Sustainable development for this area will also include stakeholder partnership, creation of ethically well-balanced datasets, use of lightweight modeling techniques, and active benchmarking, all of them under the endorsement of regional initiatives. These are needed for the creation of inclusive and context-aware ASR systems to serve the continent's diverse societies efficiently.  
\end{abstract}




\renewcommand{\shortauthors}{}

\maketitle
\pagestyle{plain}

\section{Introduction}
Human interaction is predominantly described by speech, as it is the most natural and effective modality. Automatic Speech Recognition (ASR) is applied to transform speech into written text, thereby complementing interactions between computer and humans while fostering a better understanding of semantics \cite{Yu2020, Zhou2024}. ASR has been borrowing from a variety of fields, including linguistics, computer science, acoustics, signal processing, and artificial intelligence, and has undergone remarkable strides in the past few years, being adopted in healthcare \cite{jelassi2024revolutionizing, kumar2024comprehensive}, defense \cite{hu2021artificial, guo2023towards}, and a variety of daily operations. However, low-resource languages face challenges due to the scarcity of transcribed data, linguistic resources, and, in a few cases, a scarcity of written forms of a language \cite{Zhou2024, imam-etal-2025-automatic}. Furthermore, it is estimated that more than 40\% of the world's over 7,000 languages are at risk of extinction\cite{Yu2020}.

This global challenge of developing ASR for low-resource languages is obvious in Africa, a region with massive linguistic diversity and limited digital resources. The population of Africa reached 1.29 billion in the year 2024 and continues to grow \cite{WorldBank2024}. The majority of the Africans live in rural areas and are in the profession of agriculture. In countries such as Burundi, Uganda, Tanzania, and Ethiopia, more than 60\% of the workforce are engaged in farming \cite{WPR2025}. At the same time, literacy is limited, 51.8\% in Ethiopia and 38\% in Niger, and similarly in most states \cite{WPR2025literacy}. Combined with the 1,500–2,000 languages spoken on the continent by principal language families \cite{NationsOnline2025}, these circumstances make text-based digital systems inaccessible to most of the population. Africans primarily speak in their local languages, and hence they do not use English or French locally, so automatic speech recognition (ASR) emerges as a significant research field for digital inclusion.

Despite this need, mainstream voice assistants like Siri, Alexa, and Google Assistant recognize no African languages. Early efforts like ALFFA \cite{Besacier2016} created Swahili, Hausa, Amharic, and Wolof, and the Naija Voices project is creating 500-hour corpora for Igbo, Yoruba, and Hausa \cite{LacunaFund2021}. 
The only other datasets that currently exist are BembaSpeech \cite{Chisanga2021}, ÌròyìnSpeech \cite{Adelani2024}, and Kallaama \cite{Diouf2024}. 
South Africa’s NCHLT corpus also covers the 11 official languages \cite{Barnard2014}. 
These initiatives demonstrate progress but indicate narrow scale coverage. 
By comparison, global ASR resources such as LibriSpeech (1000 hours of English audiobooks; \cite{Panayotov2015}) or Multilingual LibriSpeech (MLS; 50,000 hours in eight languages; \cite{Pratap2020}) dwarf African datasets in terms of scale, demonstrating the radical resource disparity. 
Long-standing challenges include data scarcity, tonal and morphological complexity, code-switching, dialectical variation, and limited computational resources \cite{Ndiaye2025}. Furthermore, the majority of projects still rely on text-based models that marginalize oral traditions and reproduce colonial language hierarchies \cite{Held2023}.

This systematic review investigates the gaps by synthesizing ASR research for African languages along six dimensions: (1) dataset availability and coverage, (2) data quality and licensing, (3) model architectures and training strategies, (4) evaluation metrics, (5) challenges, and (6) future directions. Our aim is to chart existing work, identify limitations, and offer guidance on creating inclusive, speech-enabled technologies for Africa's multilingual societies.

\section{Related Work}
The literature contributions in the field of ASR for African low-resource languages are still minimal. The first contribution is a challenge and future directions manuscript \cite{imam-etal-2025-automatic} that highlights only the challenges and the future directions for the African low-resource languages.  However the review lack systematic evidence collection. The subsequent work is a benchmarking and mapping survey of \cite{elmadany2025voice}, which compiles datasets, introduces SimbaBench, and assesses new models. However, its primary emphasis lies on the development of infrastructure rather than on organized evidence synthesis. In contrast to these studies, our research distinguishes itself as the first to SLR following PRISMA methodology, encompassing 71 studies conducted from 2020 to 2025. The comparative scope of these three papers is summarized by the table \ref{tab:slr_comparison}. A x means features absent from the last two papers but present in the SLR. 

\begin{table}[!h]
\centering
\small
\begin{tabular}{p{6cm}p{2cm}p{4cm}}
\hline
\textbf{Our SLR} & \textbf{\cite{imam-etal-2025-automatic}} & \textbf{\cite{elmadany2025voice}} \\
\hline
PRISMA-style systematic protocol & x & x \\
Quantitative screening scale & x & x \\
Study quality assessment & x & x \\
111 langs. across 74 dataset  & x & scope is 61 langs. \\
Benchmark/split audit & x & provides new splits, but not a field-wide audit \\
Dataset bias analysis & x & x \\
Metric critique for African languages & x & x \\
Recording conditions survey & x & focuses on benchmark composition, not condition taxonomy \\
Synthesis of models/training & Narrative, not synthesis & Evaluations, but not synthesis \\
Field-wide trends (countries, venues, years) & x & x \\
\hline
\end{tabular}
\caption{Differences between the SLR and the past review}
\label{tab:slr_comparison}
\end{table}

This review represents the first comprehensive SLR of ASR for the African low-resource languages. As opposed to narrative reviews or benchmark reports, it provides a transparent, reproducible, and comprehensive synthesis of the research covering datasets, models, metrics, and practice. Given the extreme linguistic diversity of the continent and the necessity for current digital inclusion, the review comes at an opportune time and is needed. The review identifies the necessity for ethically diversified datasets, clear benchmarking parameters, lightweight modelling, and local collaboration. As a study uncovering gaps and combining evidence, the review offers a solid base for informing future research and ensuring the development of the African ASR systems as inclusive and context-aware.

\section{Research Methodology}\label{sec2}
\begin{figure*}[ht]
\centering
\includegraphics[width=\linewidth]{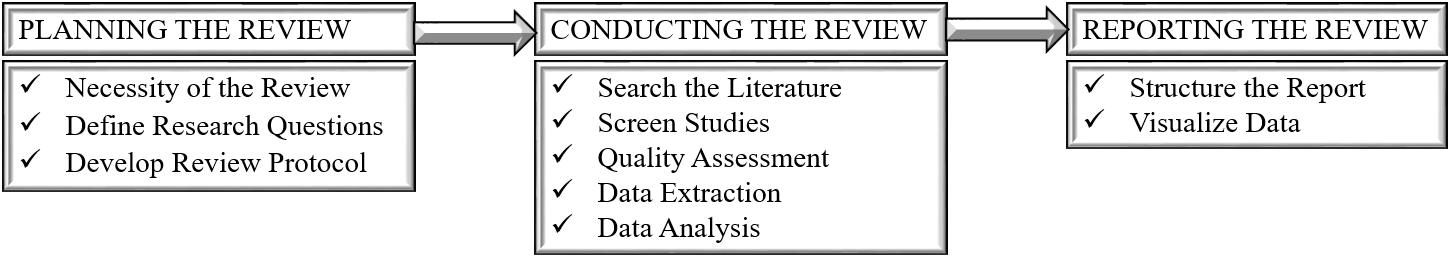}
\caption{The process of Systematic Literature Review (SLR) methodology.}
\label{figure:slr_steps}
\end{figure*}

This Systematic Literature Review (SLR) was conducted following the Preferred Reporting Items for Systematic Reviews and Meta-Analyses (PRISMA) framework \cite{page2021prisma} to ensure a transparent and replicable approach. The process depicted in Figure \ref{figure:slr_steps}, involved structured planning to address all research questions.
 
\subsection {Planning}\label{subsec2}
This phase shows the review process, which includes the necessity of the review, defining research questions, and developing a review protocol.

\subsubsection{Necessity of the Review} 
The development of ASR for low-resource African languages faces several obstacles. The purpose of this review is to conduct an in-depth analysis of datasets and models for African ASR, providing valuable insights for researchers, developers, policymakers and NLP community in general.

\subsubsection{Research Questions (RQs)}
The following research questions guide this review, pointing to key challenges in the development of ASR systems for low-resource African languages:

Bibliography Metadata:
\begin{enumerate}
    \item Which African countries are making significant contributions to the development of ASR for low-resource African languages? 
    \item How did the landscape of ASR research for low-resource languages in Africa evolve over the years?
    \item What types of venue did the authors use for ASR studies in low-resource African languages?
\end{enumerate}

Datasets:
\begin{enumerate}
    \item In what ways does dataset bias affect the performance of ASR models for African low-resource languages?
    \item What are the size and sources of the African ASR dataset?
    \item How are different dataset types distributed in African ASR research? 
    \item Do the authors provide dataset licensing, splits, and benchmark protocols in African ASR?
    \item To what extent do African speech datasets cover the linguistic diversity of the continent?
    \item What recording conditions are reported in African speech datasets?
    \item How is annotation quality addressed in African speech datasets?
    \end{enumerate}

Models:
\begin{enumerate}
    \item What ASR models are commonly used in African low-resource languages, and what are their strengths and limitations?
    \item What training techniques are generally employed to improve ASR performance in African low-resource languages?
    \item Which evaluation metrics are used to assess the ASR performance for African low-resource languages?
    \item How ASR for African low-resource languages perform in the real world?
\end{enumerate}

The RQs follow the PRISMA framework for synthesis. Bibliography Metadata: The driving force behind these questions is to reveal the wider research landscape of African ASR by determining contributor countries that are at the forefront, research over time, publication outlets of results, and field-leading keywords. The goal of dataset-oriented questions is to evaluate the foundation of African research in ASR data by examining factors like dataset size, sources, distribution, and licensing for transparency and access. We consider Africa's linguistic diversity, recording environments, and annotator quality, while also questioning how benchmarks, splits, and protocols impact model comparability and progress. Model and architecture-related questions are motivated by the need to identify the leading ASR models, their relative merits and drawbacks, and methodologies employed to cope with the problems of data scarcity. In addition, these questions probe into the operation of performance evaluation through evaluation metrics and the extension of these systems from experimental controlled settings to practical deployment in real-world settings. 

\subsubsection{SLR Protocol}
The SLR protocol provides a structure for examining ASR systems in African low-resource languages. The procedure includes getting primary research from important databases, setting up criteria for including and excluding studies, systematically extracting data, and assessing the quality of the studies.

\subsection{Conducting the SLR}\label{sec4}
Conducting the SLR is essential and starts after finalizing the protocol \cite{page2021prisma}. This section implements all the protocol steps to meet research objectives.

\subsubsection{Study Selection Criteria}
\begin{figure*}[h!]
  \centering
  \includegraphics[width=\linewidth]{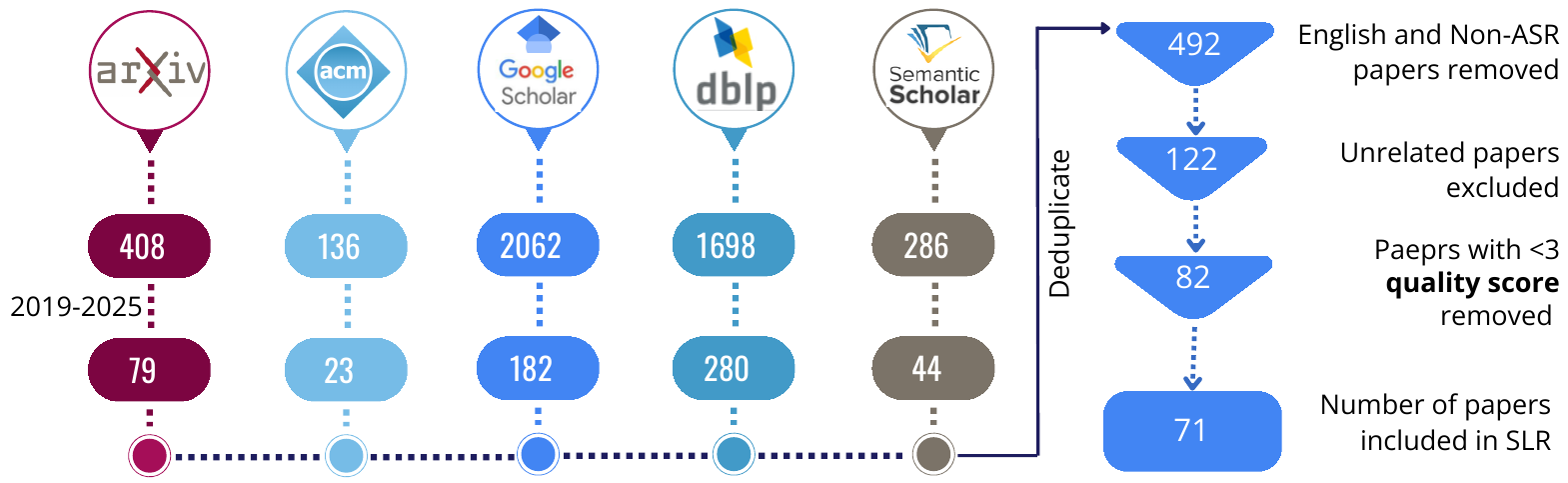}
  \caption{An overview of the ASR paper selection steps.}
  \label{fig:Paper-select}
\end{figure*}

Inclusion criteria (IC) and exclusion criteria (EC) were established to ensure the relevance of studies. IC includes papers on ASR for African low-resource languages that were published between January 2020 and July 2025, as well as datasets, models, and evaluation techniques.  EC includes non-ASR studies, studies using non-African languages, duplicate records, and studies with a quality assessment score of less than 3 out of 5. 
A structured PRISMA process was used. It start with the identification of 2,062 records, followed by the removal of 492 duplicates and the exclusion of 408 non-ASR papers during screening. At the eligibility stage, 122 unrelated papers were removed, and a further 82 studies were excluded for having quality scores below 3/5, resulting in a final inclusion of 71 studies in the review. Figure \ref{fig:Paper-select} summarizes the process.

\subsubsection{Identification of Research Works}

The identification of research began with a systematic search of the literature. A comprehensive search strategy was designed using a combination of keywords and boolean operators (AND, OR) related to ASR for African low-resource languages. The primary search strings include \texttt{automatic Speech Recognition OR ASR} AND \texttt{low-resource languages OR under-resource languages} AND \texttt {African language ORAfro-asiatic OR Niger-Congo OR Amharic OR Hausa OR Yoruba OR Igbo OR Oromo OR Swahili} AND \texttt{speech corpora OR speech corpus OR annotated speech dataset) AND models}. This query was adapted to the specific syntax of each electronic database. The review uses five databases: DBLP, ACM Digital Library, Google Scholar, Semantic Scholar, and arXiv. DBLP and ACM provide peer-reviewed papers, Google Scholar and Semantic Scholar allow larger interdisciplinary discovery, while ArXiv accommodates preprints and open-access contributions. The final data of the search from all the databases was July 1, 2025.

\subsubsection{Study Quality Assessment}

The SLR protocol provides a framework for reviewing ASR systems in African low-resource languages.  The procedure entails retrieving primary studies from relevant databases, developing inclusion and exclusion criteria for study selection, extracting systematic data, and performing quality assessments. These criteria ensure methodological accuracy and domain relevance. All papers scored a minimum of 3.5 out of 5, indicating strong quality and relevance, as shown in the \ref{tab:asr_assessment}

\begin{table}[!h]
\centering
\begin{tabular}{ll}
\hline
\textbf{SN} & \textbf{Quality Assessment Question} \\
\hline
Q1 & Does the study state its research aim related to ASR for African low-resource languages? \\ 
Q2 & Is the speech dataset well-described? \\
Q3 & Does the ASR model architecture define the task? \\
Q4 & Are ASR metrics reported and explained? \\
\hline
\end{tabular}
\caption{Quality Assessment Questions for ASR Studies}
\label{tab:asr_assessment}
\end{table}



\subsubsection{Data Extraction}
After the final selection of studies, data extraction was performed using a standardized pre-tested form in Microsoft Excel. In order to ensure accuracy, the data extraction was undertaken individually by one reviewer and then subjected to systematic checking by a second reviewer. Inconsistencies in the extracted data were discussed and cleared between the two reviewers. The extracted data encompassed multiple key domains: bibliographic metadata (such as authors, publication year, type of venue, and relevant countries), dataset characteristics (such as corpus name, languages covered, total hours of speech, conditions of recording, and license details), technical details and model specifications (including architecture and training procedures), as well as evaluation metrics and key findings. This Comprehensive extraction of such data allowed for an extensive synthesis of the field along the six dimensions outlined in the research questions.

\section{Study Characteristics}
\subsection{Bibliographic Metadata}

\subsubsection{Which African countries are making significant contributions to developing ASR for African low-resource languages?}
An analysis of empirical output as represented in figure \ref{fig:Paper-year} identifies that significant progress toward ASR development for low-resource African languages is found to be concentrated among a small number of nations. Of particular prominence is Ethiopia as a leading contributor with substantial output to Amharic, Oromo, and Tigrigna. Nigeria follows closely with a substantial amount driven by Yoruba, Hausa, and Igbo. South Africa also has a sizeable presence with a particular emphasis on its official languages, namely isiZulu, isiXhosa, and Afrikaans. There is also a substantial contribution found from work with languages most associated with Kenya, for example Swahili, as well as from the Democratic Republic of the Congo, with a particular reference to Lingala. This distribution identifies a streamlined but uneven geographical effort to offset the ASR data shortage among languages in the continent.


\subsubsection{How did the landscape of ASR research for low-resource languages in Africa evolve over the years?}
The figure \ref{fig:Paper-year} indicates the publication distribution from 2020 to 2025. In 2020, 13 publications were made but decreased to 9 in 2021 as a low point during early years. Production recovered again in 2022 with 12 publications and increased further in 2023 to 14. It reached a high point in 2024 with 17 publications as a maximum over the observed span. 2025, however, showed a dramatic decrease to 6 publications, fewer than half a year earlier. Generally, the trend indicates a consistent rise in publications between 2021 to 2024 with a dramatic sink in 2025, which is obvious as this research is carried out at the mid of the 2025.

\subsubsection{What types of venues did authors utilize for ASR studies on African low-resource languages?}
\begin{figure*}[h!]
  \centering
  \includegraphics[width=\linewidth]{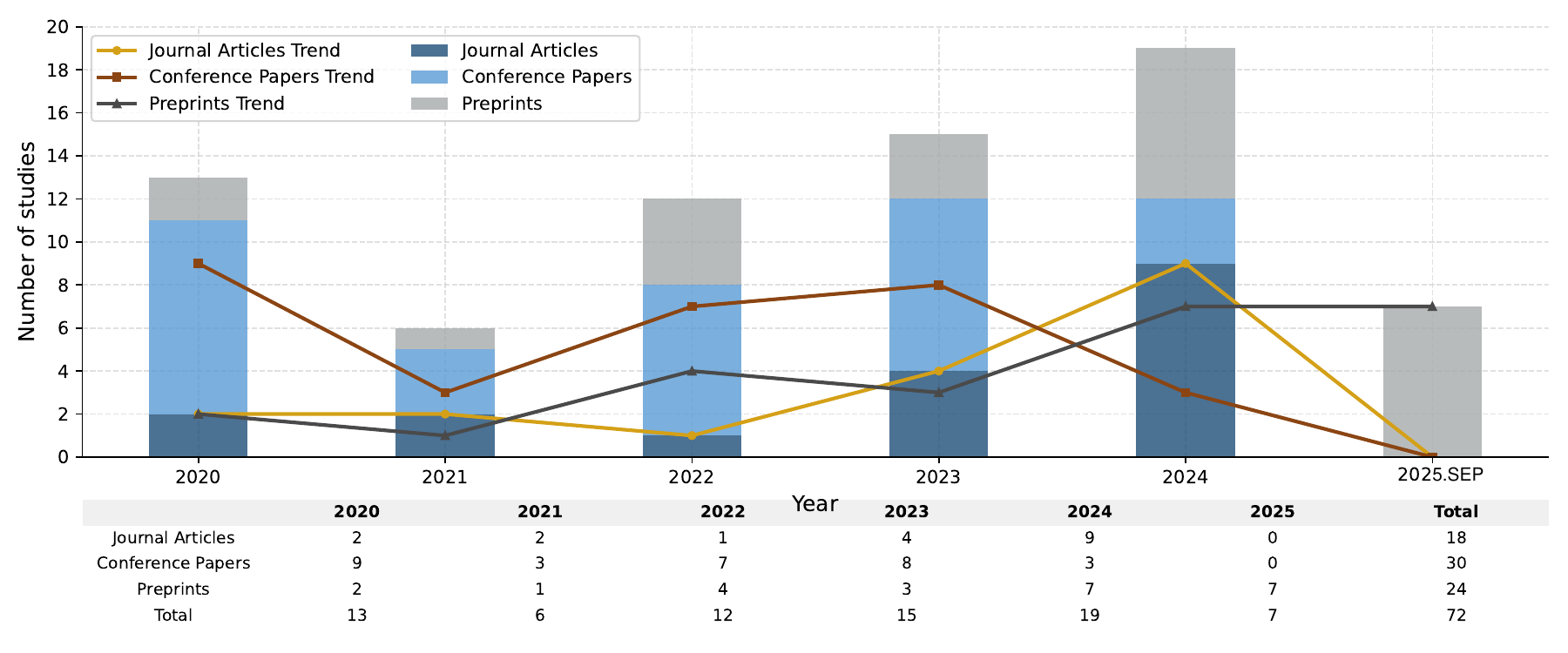}
  \caption{Distribution of ASR studies per publication years and type}
  \label{fig:Paper-year}
\end{figure*}

Based on figure \ref{fig:Paper-year}, conference papers were the most adopted venue with 40.3\% of total outcome, followed by preprints, which represent 34.7\%. Finally, journal articles with 25\% were used less frequently.

\subsection{Dataset}\label{sec:dataset}

The review covered 71 papers on ASR for African languages and compiled information on 74 distinct speech datasets encompassing 111 unique languages. The review focused on how these datasets are constructed, the amount of usable speech they provide, the sources from which the data are generated and the extent to which they are accessible to researchers and practitioners. 
\subsubsection{In what ways does dataset bias affect the performance of ASR models for African low-resource languages}
\begin{figure}[htbp]
  \centering
  \includegraphics[width=\linewidth]{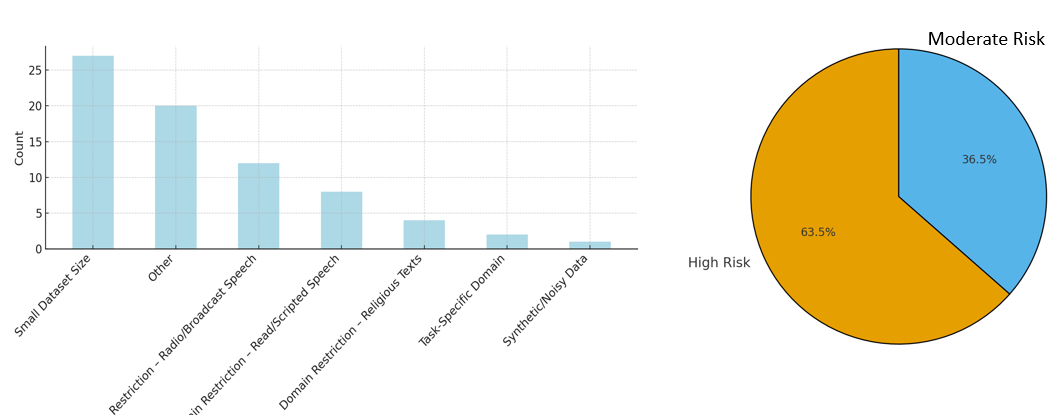}
  \caption{Datasets risks of bias}
  \label{fig:risk_of_bias}
\end{figure}

We assessed the risk of bias of 74 African ASR datasets. The Figure \ref{fig:risk_of_bias} shows that approximately 63.5\% of the datasets fall into high risk of bias and 36.5\% fall into moderate risk. The most frequent type of dataset bias is small dataset size, which undermines representativeness and generalization since corpora with only a few hours of speech or limited speakers cannot capture the linguistic and acoustic diversity of the population. The second most frequent issue is radio and broadcast restriction, which biases the data toward formal speech registers while omitting conversational and spontaneous usage. Read or scripted speech follows as another significant concern, because reliance on prepared texts or oral narratives leads to less natural linguistic variation. Religious domain corpora, although fewer in number, remain problematic due to their narrow vocabulary and repetitive pattern.
Although synthetic/noisy data sources, task-specific command corpora, and single-speaker datasets appear less frequently, they are still critical contributors to dataset bias. Overall, the bigger issue remains structural overuse of narrow domains which consistently record non-natural, non-spontaneous, and non-diverse language. This highlights the urgent need to develop larger, multi-domain, and conversational corpora to mitigate systematic bias in African language technology.

\subsubsection{How is dataset size reported in African ASR?}
\begin{figure}[htbp]
  \centering
  \includegraphics[width=\linewidth]{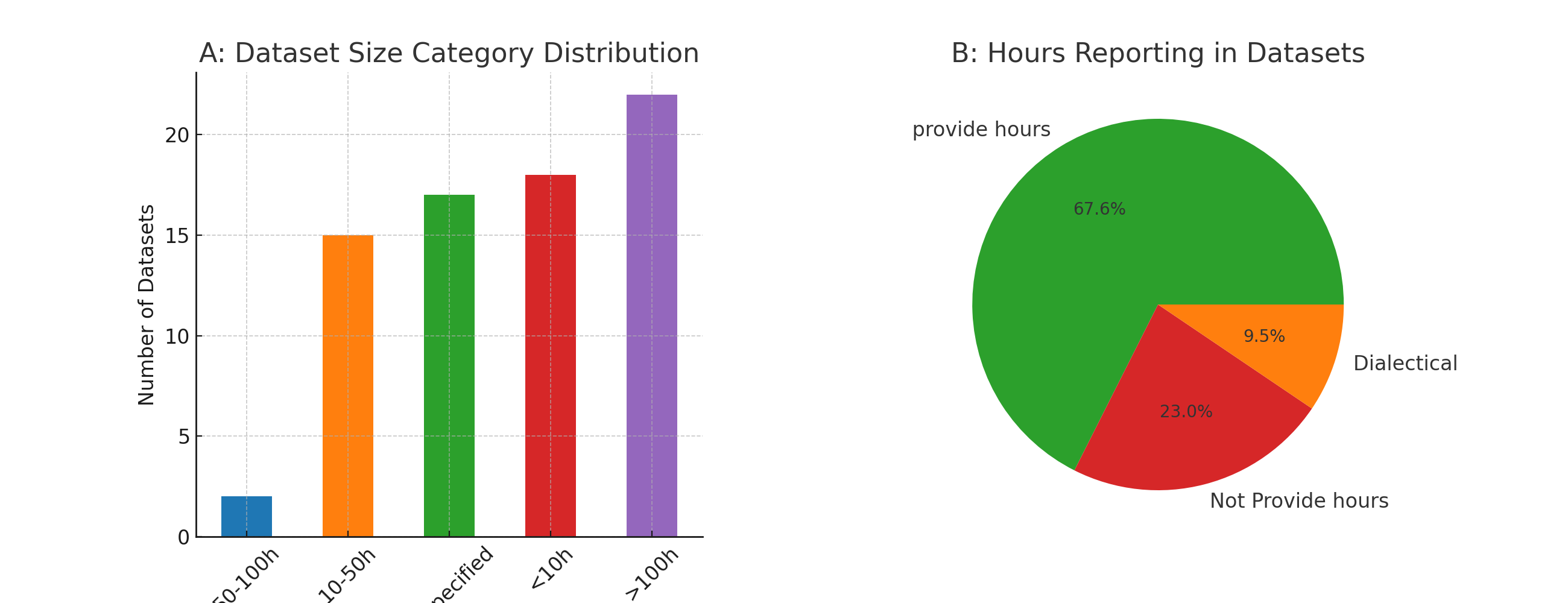}
  \caption{Distribution of Dataset Sizes and Reporting of Hours in African ASR Corpora.}
  \label{fig:dataset_size_and_hours_labeled}
\end{figure}

\begin{figure}[htbp]
  \centering
  \includegraphics[width=\linewidth]{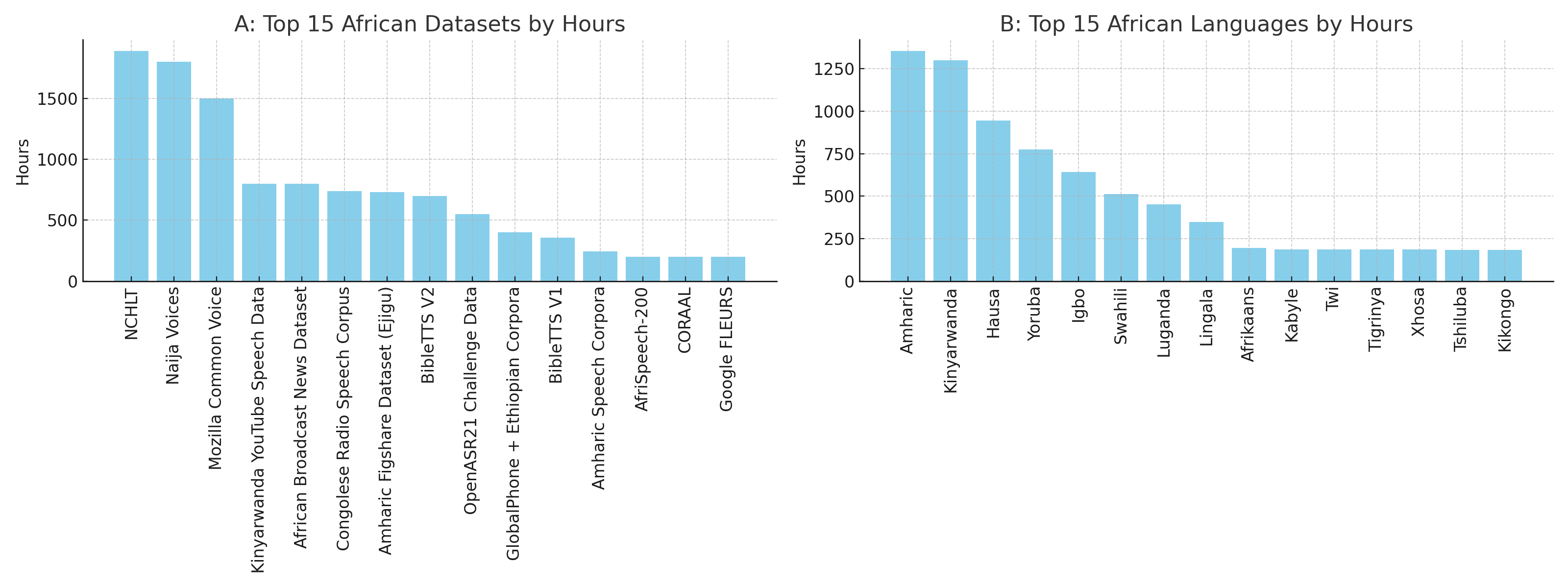}
  \caption{Size category distribution of ASR datasets.}
  \label{fig:Datasets_Languages}
\end{figure}

The total volume of African‑language speech amounts to approximately 33,197 hours. Reporting of the dataset size is inconsistent. Across the 74 datasets in, 57 datasets provided an explicit number of hours, whereas 17 datasets did not state their size. Many papers cite only the number of utterances or sentences or use vague descriptions (“several hours”), making it difficult to compare datasets (see table \ref{appendix:Dataset}). Among the datasets with reported hours, we classified them into 5 classes, see Figure \ref{fig:dataset_size_and_hours_labeled}. Remarkably, 18 datasets contain fewer than 10 hours, and only 17 exceed 100 hours. 
The Figure \ref{fig:Datasets_Languages} illustrates the top 15 African ASR datasets by speech hours, exhibiting a wildly uneven distribution of size. The Kinyarwanda YouTube Speech Data drastically overwhelms with more than 22,000 hours, yet a large proportion of this material goes untapped because it is not labeled, thus hindering it for ASR development purposes. After this corpus is excluded, the combined African-language hours for the remainder of the datasets drop to a mere 10,000 hours, thinly dispersed across a multitude of languages. By contrast, a single English dataset such as LibriHeavy offers 55,000 labeled hours, underscoring how African resources are both smaller and disrupted. This imbalance highlights a pressing challenge while English benefits from massive as well as curated corpora, African languages remain critically underrepresented, disrupting the development of robust and inclusive ASR systems.
\subsubsection{What are the source categories of African ASR dataset?}
\begin{figure}[htbp]
  \centering
  \includegraphics[width=\linewidth]{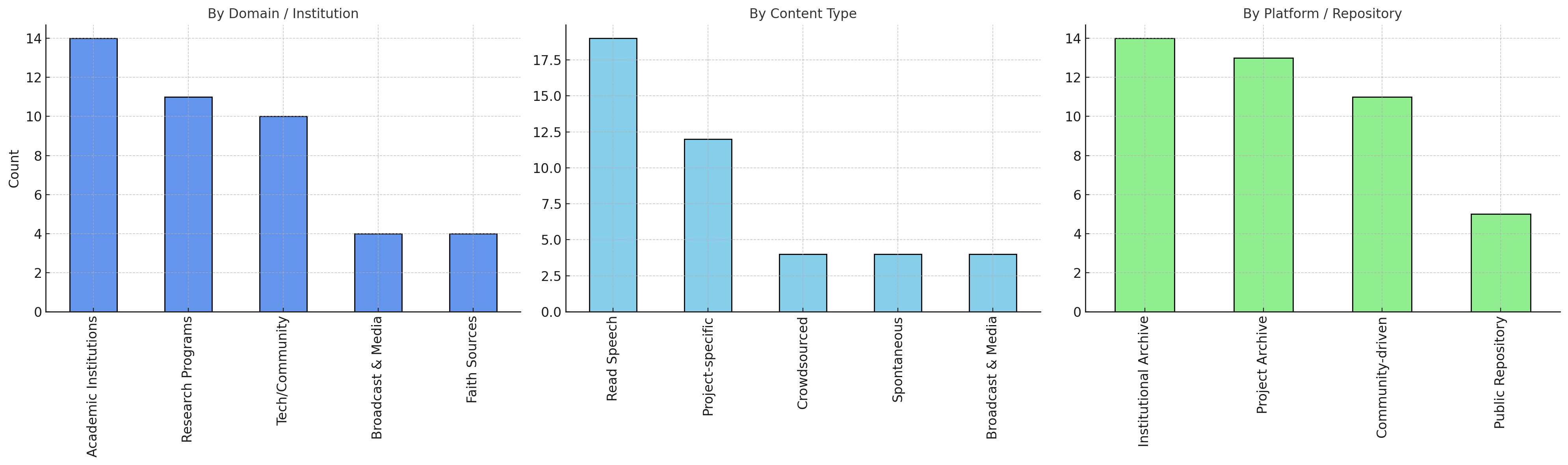}
  \caption{Classification of ASR Dataset Sources by Domain, Content Type, and Distribution Platform.}
  \label{fig:dataset_source}
\end{figure}

As seen in Figure \ref{fig:dataset_source} we tried to classify the sources into three main classes: by domain/institution, by the content they provide, and by the distribution platform/repository. The results show that the largest share of datasets comes from academic institutions, primarily contributing read speech through institutional archives, while technology companies and community platforms also play a significant role in supplying crowdsourced and spontaneous speech. Research projects and campaigns are a major contributor, often producing datasets unique to specific projects. In terms of types of content, read speech dominates, indicating strong academic and systematic collecting procedures; however, spontaneous, broadcast, and crowdsourced speech are relatively undercovered and therefore show a scarcity of natural conversation-type data. In terms of distribution platforms, datasets are mostly archived within institutional collections or distributed via community-driven repository sites, with lesser, but still present, contributions from project-local repository sites and public repository sites. This classification underscores the centrality of academia and research initiatives in dataset creation while pointing to the need for broader diversification of speech types and open distribution mechanisms.
\subsubsection{How are different dataset types distributed in African ASR research?}
\begin{figure}[htbp]
  \centering
  \includegraphics[width=\linewidth]{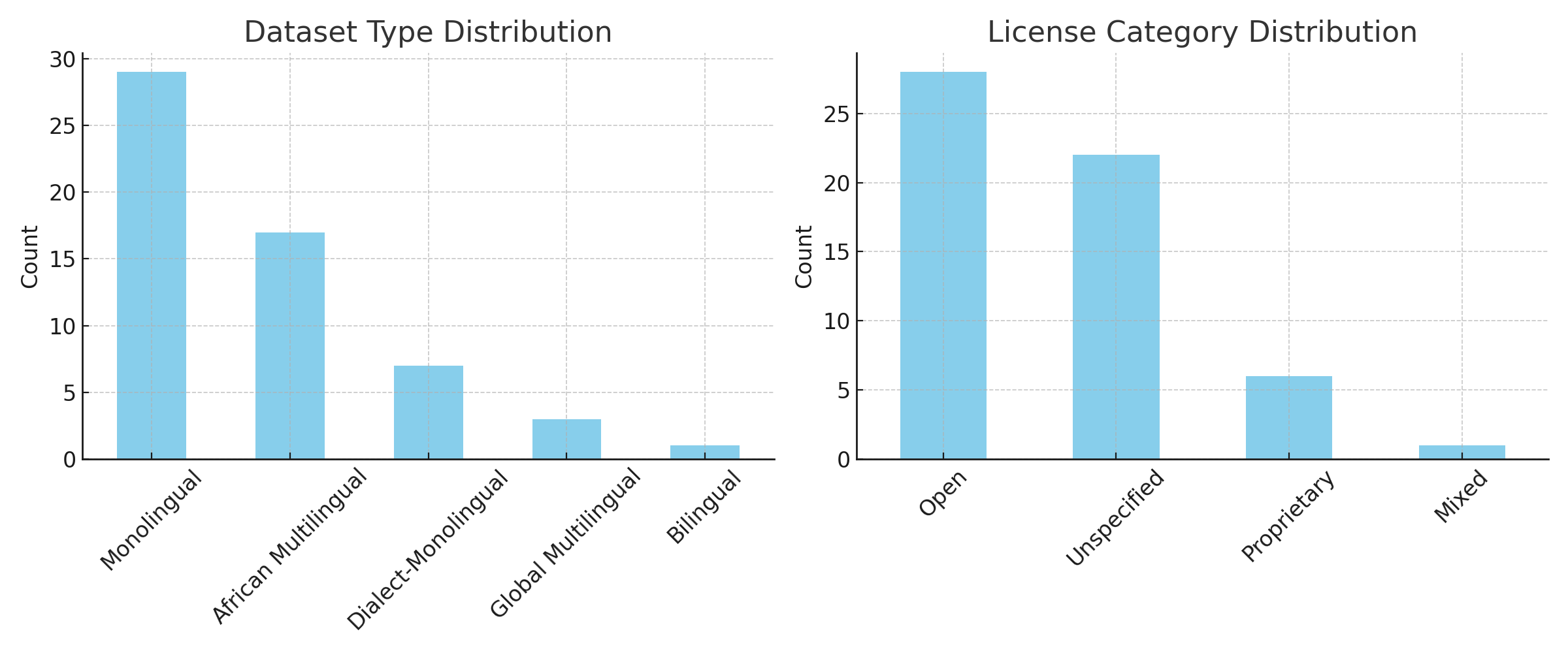}
  \caption{Source category distribution of ASR datasets.}
  \label{fig:Dataset_type_Licence}
\end{figure}

The Figure \ref{fig:Dataset_type_Licence} shows the distribution of dataset types in African ASR research. Most of the datasets are monolingual (34), dealing with one specific African language, reflecting the importance of recording individual languages. The second most numerous category is African multilingual datasets (21), bringing together several African languages in one resource. Dialectal monolingual datasets (11) deal with English speech using African accents and serve to develop ASR for accented English speech. Global multilingual datasets (4) contain African languages, as well as numerous world languages, and offer wider coverage, but narrow African-specific content. Few code-mixing datasets (3) reproduce language switching, which reflects practical communication in life, while just one can be found of bilingual datasets, serving just two languages. The statistics demonstrate monolingual corpora prevalence, but also the increasing significance of multilingual, as well as dialectal, datasets in creating an inclusive ASR system.
\subsubsection{Do the authors provide dataset licensing, splits, and benchmark protocols in African ASR?}
The Figure \ref{fig:Dataset_type_Licence} shows that most datasets are openly licensed, supporting accessibility and collaboration, while many others have licenses that are not specified in the papers, creating uncertainty for reuse. Proprietary and mixed licenses are fewer, reflecting restricted availability but also highlighting the need for clearer reporting of dataset licensing.
In ASR, having access to well-licensed, reproducibly split, and benchmark-ready datasets is a necessity. For African languages datasets such as Mozilla Common Voice under a CC0 license is an example of public dataset. This license allows unrestricted use and reproducing making it perfect for both research and commercial applications. Additionally, the Voice of Naija under CC BY-NC-SA 4.0, present a large-scale, corpus for non-commercial consumption.
\cite{peterson2022openasr21} applies blind evaluation processes, providing fixed BUILD, DEV, and EVAL sets across several languages. These ensure that evaluations are fair and meaningful across diverse systems. In the same vein, \cite{biswas2022code} proposes multilingual code-switching benchmarks, advancing the complexity of evaluation beyond monolingual WER. Benchmarks like these not only guide model development but also shape the research community’s focus toward real-world complexity, multilingualism, and inclusivity. Table \ref{appendix:licenses} from the appendix provide the details.
Licensing is important, also, ensuring clear dataset splits is essential for reasonable testing and consistent evaluations in research. This helps researchers run repeatable experiments and establish trustworthy benchmarks. It ensures that performance measures, like WER, are accurate and not inflated by overlapping data. Contrary to this, their usefulness is limited for community tests and benchmarks. In a field that values accuracy, reproducibility must be a priority.

\subsubsection{To what extent do African speech datasets cover the linguistic diversity of the continent?}
In terms of language coverage, the review identified 111 African languages represented across the surveyed datasets. Of these, 97 languages had an estimated number of available speech hours, while the remaining 14 did not provide any explicit information on corpus size. As shown in the figure \ref{fig:dataset_source}, the distribution of speech hours is highly inconsistent, with Kinyarwanda dominates the coverage, with more than 22,000 hours of unlabelled YouTube dataset, while Amharic follows as the second most resourced language. Other languages such as Hausa, Yoruba, Igbo, Swahili, and Luganda appear with modest but still notable amounts of data see Figure \ref{fig:Datasets_Languages}. The long tail of the distribution reflects the reality that the majority of African languages have only limited or no quantified resources, underscoring the imbalance in data availability and the urgent need for the development of well-labeled dataset across the African low-resource languages.

The diversity of African languages presents both opportunities and challenges for the development of ASR for African low-resource languages.
Amharic is represented in over 15 datasets \cite{tachbelie2020dnn, demisse2024amharic, ejigu2024large, asfaw}, with some datasets including dialectal disparities such as Gojjam, Wollo, and Shewa. Likewise, Yorùbá is included in at least 9 datasets, with substantial contributions from \cite{emezue2025naijavoices} large-scale corpus and \cite{gutkin2020developing} fully diacritized recordings. These contributions serve as a way for ASR development, orthographic modeling, and diacritic restoration in tonal languages.
Moreover, Hausa is represented in 7 different datasets. \cite{emezue2025naijavoices} insertion of Hausa is impactful. Notably, \cite{abubakar2024development} emphasizes that preserving diacritic and tonal markings in Latin script is an essential feature for high-quality linguistic processing. This corpora, together with the contributions of \cite{ibrahim2022development, meyer2022bibletts}, sets a solid foundation for high-quality Hausa ASR systems. 
Numerous African languages, like Dagbani, Fon, Bambara, and Tshiluba, are underrepresented in available datasets, often appearing only once in the collections. However, this offers a chance for further research and marginalized languages will be included. Table \ref{appendix:Linguistic} in the appendix provides a detailed explanation.
To create an effective dataset, collective efforts must put together beyond just language classification. The best datasets today, such as those by \cite{emezue2025naijavoices, nzeyimana2023kinspeak, afonja2024performant}, are not just large but are inclusive, multi-accented, and built for reproducibility. Their success shows that with community engagement, creative augmentation, and open licensing, the creation of a large dataset can scale language resources without sacrificing quality.

\subsubsection{What Recording Conditions are reported in African Speech Datasets?}
The recording environment is very important in determining the performance of ASR in real-world. Datasets from \cite{tachbelie2020dnn, demisse2024amharic} use a controlled environment to record high-quality audio. They reduced noise with quality microphones like the \textit{Shure SM58} and pop filters, ensuring correct phoneme-level alignment and transcription. This methodology promotes effective benchmarking and academic research.
Models trained exclusively on clear studio recordings consistently face difficulties when applied to noisy or real-world. To solve this issue, \cite{emezue2025naijavoices} used mobile applications and various speaker locations to replicate authentic usage settings. Likewise, \cite{nzeyimana2023kinspeak} built a corpus of YouTube talk programs, thereby integrating acoustic variability and uncontrolled speaker settings. While this method has setbacks, such as overlapping voice and background music, it significantly advances the dataset's environmental validity, which cannot be duplicated in any studio setup. 
Hybrid datasets, which combine high-quality speech with diverse content, such as broadcasts, are beneficial in improving ASR systems. Researchers like \cite{asfaw, ejigu2024large} improve their Amharic dataset with features like specification and noise variations to preserve transcription quality under realistic settings. Also, \cite{westhuizen2021multilingual, peterson2022openasr21} include recordings from radio news and interviews, representing real-world interruptions. This diversity in data is essential for enhancing ASR performance in places such as homes and clinics.
The greatest speech datasets do not necessarily include clean audio; they should include ambient noise. This necessitates a range of languages, microphones, settings, accents, and background noises. \cite{emezue2025naijavoices, afonja2024performant} prove that, with community support, it is possible to build large, high-quality datasets in real-world settings.

\subsubsection{How is Annotation Quality addressed in African speech datasets?}
Accurate annotations are critical in ASR, particularly for African languages with complex tonal systems, diacritics, or language switching.  Most of the datasets, such as \cite{tachbelie2020dnn, demisse2024amharic}, display high annotation quality.  They offer manually checked transcriptions, phoneme- and grapheme-level alignments, and evaluation sets featuring various speakers.  These allow the researchers to analyse CER and WER, as well as understand how excellently the ASR operates and how accurately it reflects the language.
 \cite{emezue2025naijavoices} performs double human review on transcriptions, and the dataset offers JSON manifest files with rich metadata for repeatability and downstream benchmarking.  Similarly, \cite{abubakar2024development} contributes further by specifically evaluating diacritic alignment in their Hausa corpus using DER and WER.  This is especially important in tonal languages, where missing diacritics can totally alter word meaning.  
Regardless of their size, large corpora like \cite{nzeyimana2023kinspeak} are unlabelled and lack standardized annotation techniques. Others, such as \cite{ritchie2022large} combined a multilingual set, draw from diverse sources with irregular annotation systems, preventing cross-language comparison.  While corpora like \cite{biswas2022code} include unique features such as word-level language-switch tags, code-switched speech remains the exception rather than the usual, as the majority of corpora still lack standardized annotation techniques. This disparity restricts their immediate usefulness for model training and evaluation, particularly in scenarios that need multilingual or dialect-aware systems.
It is vital to understand that high-quality annotation is not an option, but rather a need for fair, accurate, and scalable African speech technology. Consistent, linguistically aware annotation is essential for training models that accurately represent the intended voices. Table \ref{appendix:annotation quality} in the appendix contains the detail.

\subsection{Model Architicture}\label{sec:model}
\subsubsection{What ASR models are commonly used in African low-resource languages and what are their strengths and limitations?}

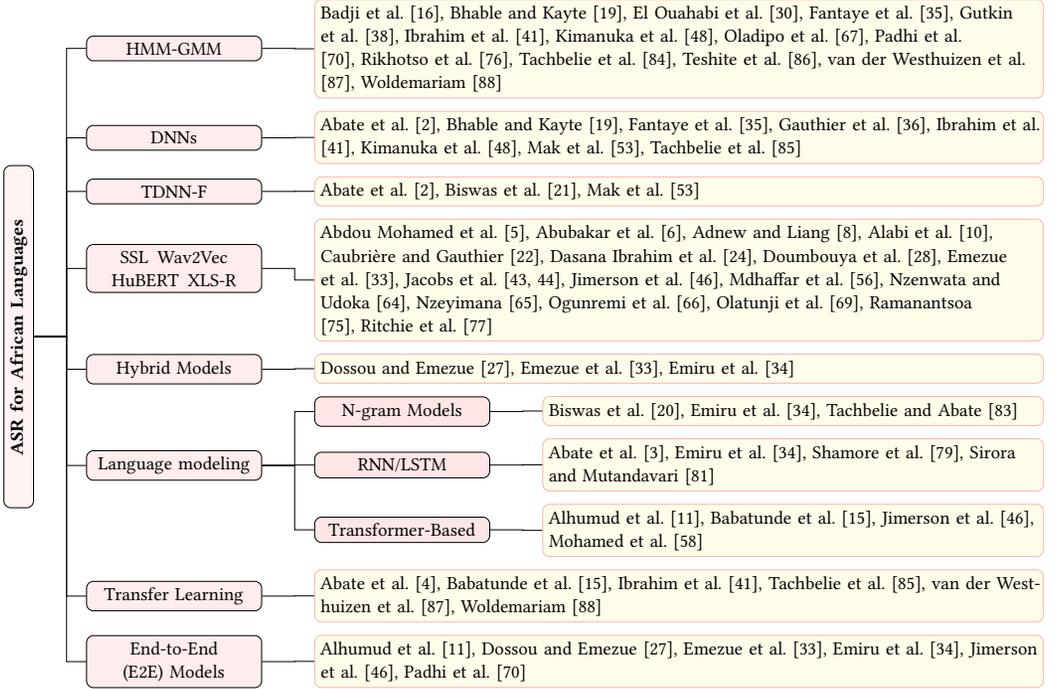
\begin{figure}[h!]
    \centering
    \footnotesize
\tikzset{
    rootbasic/.style = {draw, text width=5cm, rounded corners=3pt, align=center, rectangle, rotate=90, font=\footnotesize, fill=pink!20},
    level1/.style = {draw, text width=2.5cm, rounded corners=3pt, align=center, rectangle, font=\footnotesize, fill=pink!30},
    level2/.style = {draw, text width=2.5cm, rounded corners=3pt, align=center, rectangle, font=\footnotesize, fill=pink!40},
    level3/.style = {draw=red!30, text width=11cm, rounded corners=3pt, align=left, rectangle, font=\footnotesize, fill=yellow!10},
    level3narrow/.style = {draw=red!30, text width=7.5cm, rounded corners=3pt, align=left, rectangle, font=\footnotesize, fill=yellow!10},
}
\resizebox{\textwidth}{!}{%

\begin{forest} 
for tree={  
    grow'=east,
    growth parent anchor=east,
    parent anchor=east,
    child anchor=west,
    calign=center,
    l sep=8mm,
    edge path={
        \noexpand\path[\forestoption{edge}]
        (!u.parent anchor) -- +(5mm,0) |- (.child anchor)\forestoption{edge label};
    },
}
[\textbf{ASR for African Languages}, rootbasic, parent anchor=south,
    [HMM-GMM, level1, yshift=-5.4mm,    [\citet{tachbelie2020analysis,teshite2023afan,badji2020automatic,westhuizen2021multilingual,gutkin2020developing,kimanuka2024speech,ibrahim2022development,rikhotso2021development, el2023comparative,fantaye2020investigation,oladipo2020accent,bhable2020review,woldemariam2020transfer,padhi2020multilingual}, level3]
    ]
    [DNNs, level1, yshift=-2mm,[\citet{abate2020deep,fantaye2020investigation,kimanuka2024speech,ibrahim2022development,mak2024exploring,gauthier2016collecting,bhable2020review,tachbelie2020dnn}, level3]
    ]
    [TDNN-F, level1, yshift=-0.2mm,  
    [\citet{abate2020deep,mak2024exploring,biswas2022code}, level3]
    ]
    [SSL\, Wav2Vec\, HuBERT\, XLS-R, level1, yshift=-3.7mm,
    [\citet{alabi2024afrihubert,caubriere2024africa, nzeyimana2023kinspeak,emezue2023afrodigits,jacobs2023towards,ramanantsoa2023voxmg,olatunji2023afrispeech,ibrahim2023breaking,nzenwata2024ika,abubakar2024development,ogunremi2023r,jimerson2023unhelpful,adnew2024semantically,doumbouya2021using, abdou2024multilingual,ritchie2022large, mdhaffar-etal-2024-performance, jacobs2025speech}, level3]
    ]
    [Hybrid Models, level1, 
    [\citet{dossou2021okwugbe,emezue2023afrodigits,emiru2021improving}, level3]
    ]
    [Language modeling, level1,
        [N-gram Models, level2, 
            [\citet{biswas2019improved,emiru2021improving,tachbelie2023lexical}, level3narrow]
        ]
        [RNN/LSTM \\ , level2, yshift=-1.9mm,
            [\citet{abate2020multilingual,sirora2024shona,emiru2021improving,shamore2023hadiyyissa}, level3narrow]
       ]
        [Transformer-Based, level2, yshift=-1.9mm,
            [\citet{alhumud2024improving,jimerson2023unhelpful,babatunde2023automatic,mohamed2023multilingual}, level3narrow]
       ]
    ]
    [Transfer Learning, level1, yshift=-1.7mm,
        [\citet{tachbelie2020dnn,woldemariam2020transfer,abate2021endtoend,westhuizen2021multilingual,babatunde2023automatic,ibrahim2022development}, level3]
    ]  
    [End-to-End (E2E) Models, level1,
        [\citet{dossou2021okwugbe,emezue2023afrodigits,emiru2021improving,alhumud2024improving,jimerson2023unhelpful,padhi2020multilingual}, level3]
    ] 
]
\end{forest}
}
\caption{Research Taxonomy of ASR for African low-resource languages Models.}
\label{fig:model_compression_taxonomy}
\end{figure}

\begin{figure*}[t]
  \includegraphics[width=0.48\linewidth]{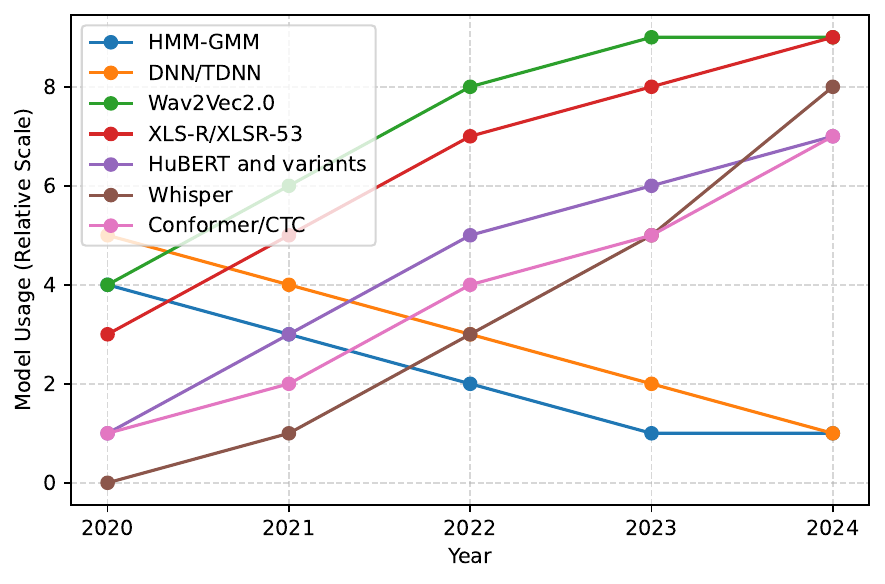}
  \includegraphics[width=0.48\linewidth]{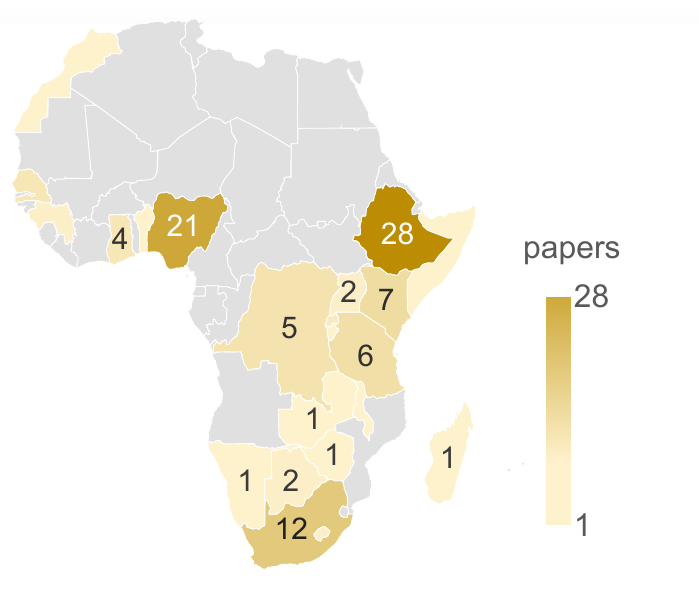} \hfill
  \caption {Model usage  trends (the left line graph) and publication statistics across countries (the right side map).}
\label{fig:Model-trend}
\end{figure*}

~\\ Finding out which ASR model types are most frequently used gives insight into current research trends.  Figure \ref{fig:model_compression_taxonomy} shows the ASR works for those focused on African languages, carefully grouped based on the methods used during the model development. Figure \ref{fig:Model-trend} reveals that the recent development in African ASR is led by SSL. Models like wav2vec 2.0, XLSR-53, and MMS show outstanding performance in multiple research. These models lead the research areas, appearing in over 17 papers due to their capability to exploit large quantities of unlabelled speech for pretraining. For instance, \cite{emezue2025naijavoices} reports a 75.86\% WER improvement using Whisper, and \cite{nzenwata2024ika} show meaningful few-shot gains on Ika with as little as one hour of Bible data. However, their reliance on GPU and sensitivity to extremely small training corpora requires careful attention.

Similarly, models like CNNs, RNNs, and HMM-GMM hybrids continue to serve critical functions, particularly where low-latency inference and minimal compute budgets are essential. These models appear in 7–10 studies, such as \cite{ayall2024amharic, rikhotso2021development}, showing consistent performance in domain-specific and limited-vocabulary tasks. Similarly, Whisper used in 10 studies, has emerged as the most robust solution for accented and multilingual African speech. However, Whisper and wav2vec both carry risks as their heavy memory requirement restrict their deployment on edge devices. Also, their inflexible architectures serve as a barrier for phoneme-level tasks like diacritic modeling. 
Studies like \cite{tachbelie2023lexical} on morpheme-based modeling in Amharic, or \cite{alhumud2024improving} on accent-aware voice conversion, prove that careful architectural and linguistic adaptation can yield WER reductions of 5–70\%, even under severe data scarcity. Yet, caution is critical because wav2vec tends to overfit without data augmentation, Whisper’s fitness constraints hinder transparency, and HMM-GMM models, though lightweight are outdated for general-purpose ASR.

\subsubsection{What training techniques are generally employed to improve ASR performance in African low-resource languages?}
Training strategies have a substantial impact on ASR performance in African low-resource languages. Researchers have used a different types of methodologies, including SSL, multilingual transfer, data augmentation, morpheme-level tokenisation, and curriculum learning. The table \ref{appendix:Training_Tech} organises these strategies, evaluating their functionalities and associated trade-offs. This provides insight into the techniques most suitable for the linguistic, acoustic, and technical contexts of African speech. ASR performance is particularly enhanced through SSL and fine-tuning, especially when combined with additional data, semi-SL, multilingual training, and data augmentation.  

\subsubsection{Which evaluation metrics are used to assess the ASR performance for African low-resource languages?}
This is an essential aspect of ASR research, particularly for African languages, where linguistic diversity, dialectal variation, and data availability present unique challenges. To accurately assess ASR effectiveness, variety of metrics, like WER, CER, DER, Medical WER, and BLEU are adopted by the researcher. 
From the table \ref{appendix:asr_metrics}, WER remains the most dominant evaluation metric across African ASR research. The challenges of processing morphologically rich tonal languages are becoming more noticeable. While metrics such as CER and DER are essential for detecting linguistic errors in such languages, they remain underutilized in ASR evaluations. Similarly, task-specific metrics such as MWER and BLEU in speech translation emphasize the importance of moving beyond traditional techniques, particularly in practical areas like healthcare and multilingual communication. 
An evaluation of ASR models for African low-resource languages demonstrates the critical balance between model complexity and language fit. Table \ref{ranking} shows that Model combination between CNN \& RNN or TL-DNNs often perform better then large language models when the dataset is clean. For examples, Amharic archieved very lowe WER (2.0\%) with these techniques \cite{ejigu2024large} and \cite{westhuizen2021multilingual} TL-DNN with WER of only 5.95\%. While large models like XLSR and MMS gave much worse results (up to 53\%) when data was imbalanced \cite{sikasote2022bembaspeech, nzenwata2024ika}. This clearly indicate that the success depends on the quality of the dataset and language-specific fine-tuning. Also, the performance varies within models, which shows the impact of data quality, dialect diversity, augmentation and training techniques. Studies using accent-aware adaptation and morpheme-based modelling improve robustness, especially in languages with complex structures. 

\begin{table}[!h]
\centering
\begin{tabular}{p{1cm}p{5cm}p{2cm}p{3cm}}
\hline
\textbf{Rank} & \textbf{Model Type} & \textbf{Av. WER (\%)} & \textbf{Authors} \\
\hline
1 & CNN+RNN & 2.00 & \cite{ejigu2024large} \\
2 & DNN + Transfer & 5.95 & \cite{westhuizen2021multilingual} \\
3 & Conformer & 9.55 & \cite{nzeyimana2023kinspeak} \\
4 & HMM & 16.50 & \cite{teshite2023afan} \\
5 & wav2vec 2.0 & 18.61 & \cite{abubakar2024development, ramanantsoa2023voxmg} \\
6 & CMU Sphinx & 18.87 & \cite{awino2022phonemic} \\
7 & MFCC + RNN & 19.30 & \cite{asfaw}\\
8 & wav2vec 2.0 + Seq2Seq & 23.30 & \cite{adnew2024semantically} \\
9 & Transformer-based E2E & 23.80 & \cite{ogunremi2023r}\\
10 & DNN Transfer & 24.50 & \cite{woldemariam2020transfer} \\
11 & CNN + LSTM & 29.00 & \cite{sirora2024shona}\\
12 & HMM-DNN & 32.80 & \cite{tachbelie2020analysis}\\
13 & XLSR & 32.91 & \cite{sikasote2022bembaspeech}\\
14 & MMS & 53.77 & \cite{nzenwata2024ika}\\
\hline
\end{tabular}
\caption{Ranking of Model Types by Average WER and Representative Studies}
\label{ranking}
\end{table}

\subsubsection{How ASR for African low-resource languages perform in the real world?}
In the development of ASR, adaptability and robustness are essential for real-world applicability. African language known by dialectal variation \cite{asfaw, oladipo2020accent, olatunji2023afrinames}, code-switching \cite{masekwameng2020effects, woldemariam2020transfer, westhuizen2021multilingual, fantaye2020investigation}, noisy environments \cite{mukiibi2022makerere, jacobs2023towards, nzeyimana2023kinspeak}, and accent diversity \cite{alhumud2024improving, olatunji2023afrinames}, all of which challenge model generalization.  To address these, researchers have employed different techniques, see table \ref{appendix:asr_strategies}. 

\section{Challenges and Future Directions}

This section discusses the key ASR challenges, highlight trends and future directions of the review in ASR for African low-resource languages. 

\subsection{Data Disparities and Dialectal Representation}
Dataset scarcity and narrow coverage remain the most frequently cited limitations, with numerous studies calling for expansion in terms of speaker diversity, dialectal representation, domain coverage, and recording conditions \cite{emezue2023afrodigits, ibrahim2022development, olatunji2023afrispeech, ogunremi2023r, meyer2022bibletts, doumbouya2021using}. A significant challenge is the imbalance in the corpus size and coverage among African languages. While languages such as Amharic, Hausa, Yorùbá, and Kinyarwanda have relatively substantial and diverse datasets, most African languages remain severely under-resourced. For example, Dagbani, Tshiluba, and Bambara often have fewer than 10 hours of usable audio. These imbalances make it difficult to build a generalisable ASR system without relying on transfer learning, data augmentation, or cross-lingual modelling. In most of the studies, dialectal variation is underrepresented, undermining fairness and usability in multilingual contexts. To tackle these challenges, future research must focus on collecting data that posses better transcription accuracy and is culturally relevant. Furthermore, it’s essential to create domain-specific datasets for areas like education, sports, and healthcare, among others.

 \subsection{Annotation Quality, Licensing, and Ethical concerns}
Benchmarking and reproducibility remain pressing challenges, underscoring the need for standardized metrics, domain-specific evaluation sets, and open-source evaluation scripts \cite{alabi2024afrihubert, jimerson2023unhelpful, olatunji2023afrispeech}. Also, the need for bias and fairness evaluation, especially in accented and African American Language recognition, to prevent biases in domain-specific datasets such as healthcare \cite{martin2023bias, babatunde2023automatic, morris2021one}. Dataset quality and accessibilty present additional barriers. High-quality datasets typically include Phoneme-level alignments, verified transcriptions, and clear evaluation splits, which support reproducibility. However, many datasets are inconsistently annotated or lack reliable labels, reducing their re-usability. Licensing also presents challenges. While open-licenses datasets have accelerated research, many others are either unlicensed or restricted to academic use, limiting it commercial application and reusability. Also, ethical issues are additional challenges as must of the datasets fail to account for gender balance, age diversity, or speaker consent. Community participation in data collection remains rare. Future research should focus on establishing standardized benchmarks, domain-specific datasets and publicly available sources to support reproducibility. Developing a dataset with reliable annotations that reflect gender, age, and linguistic diversity equally is important.

\subsection{Model Language Mismatch}
 There is also a clear mismatch between the design of current ASR models and the linguistic realities of African Languages. Many belongs to Niger-Congo and Afroasiatic families, which are tonal, agglutinative, or morphologically complex. However, most ASR models depend on subword or character-level tokenisation, which does not adequately capture these features. The absence of morpheme-level modelling, phoneme-aware architectures, and diacritic restoration leads to poor generalisation and semantic error, particularly in tonal languages. Also, some progress has been made, for instance \cite{abubakar2024development}, used DER to test tonal accuracy in Hausa, and \cite{emiru2021improving}, used morpheme-level CTC training to improve Amharic ASR, such linguistic approaches remain underutilized.

 \subsection{Evaluation Metrics}
 Orthographic and Phonological challenges, such as diacritics and inconsistent spelling make both training and testing very difficult. Althoug WER remains the dominant evaluation metric in ASR for African low-resource languages due to its simplicity and comparability, its limitations in capturing tonal errors, diacritic variation, and morphological richness reduce its usefulness for many African languages. Alternative measures such as CER and DER are rarely used, not only because of researcher preference but also due to the lack of phoneme or diacritic-level annotations in most corpora. Likewise, task-specific metrics like MWER and BLEU, which are vital for applied domains such as healthcare or translation, are underutilised since much African ASR research still focuses on basic transcription tasks. Without broader adoption of linguistically and contextually relevant metrics, performance claims risk being misleading, as systems with low WER may still fail in real-world applications. Researchers suggest using special pronunciation dictionaries and sub-word tokenization that fit agglutinative and tonal languages  \cite{nzeyimana2023kinspeak, koffi2020tutorial, dossou2021okwugbe, adnew2024semantically}. 

 \subsection{Pretrained Models and Adaptation}
 Recent progress has been driven by large pretrained models such as wav2vec 2.0, Whisper, and MMS, which show strong performance in African low-resource languages. However, without domain adaptation, these models risk creating unrealistic performance expectations. Many show biases when applied to African languages, especially in handling accents, dialects and noisy speech. Fine-tuning, therefore, is essential to ensure real-world usability.
 
\subsection{Deployment and Robustness}
There is constant consideration to robustness under real-world conditions, such as noisy or broadcast audio, and spontaneous speech inconsistency \cite{jacobs2023towards, mukiibi2022makerere, zellou2024linguistic}. Many ASR systems evaluated in controlled environments using studio-quality audio, which limits insight into their real-world robustness. As a result, their performance in noisy or variable acoustic conditions remains uncertain. Efforts such as \cite{nzeyimana2023kinspeak, mukiibi2022makerere} introduce realistic noise variability, but these datasets remain underused in benchmarking. Datasets such as \cite{afonja2024performant, peterson2022openasr21} have pioneered speaker-independent testing and blind evaluation, yet such practices are still rare. Improvements here require domain-specific noise profiles, de-noising pipelines, and explicit out-of-domain evaluation protocols.

Addressing these challenges requires collaborative effort from stakeholders, including Governments, Universities, and Local Communities, among others. By prioritising linguistically aware model design, ethically grounded data collection, and robust evaluation practices, the ASR for AFrican low-resource languages can move forward, more accurate, inclusive, and sustainable.

\section{Conclusion}
This SLR reviewed 71 studies on ASR for African low-resource languages, highlighted advancements in self-supervised learning, transfer learning, and community-driven dataset initiatives, and also revealed a shortage in dataset coverage, annotation quality, and linguistic diversity.  The existing dataset is limited by corpus sizes, inadequate dialectal representation, which reduced reproducibility and disrupted thorough model assessment.  Moreover, ethical concerns remain inadequately explored. These findings indicated that substantial inconsistency, which hindered equal performance across the different African languages.  In the absence bias assessment, linguistically informed metrics, and sustainable dataset governance, ASR systems may risk marginalisation in critical sectors such as healthcare, education, sport, etc. Subsequent efforts should focus on developing extensive, ethically curated, and openly licensed corpora implementing linguistically informed modelling techniques that accommodate tonal and morphologically complex languages and include standards and fairness assessments.  These objectives necessitated collaboration among governments, academics, industry, and the community to ensure that ASR technologies are inclusive and replicable for developing inclusive ASR for African low-resource languages.

\bibliographystyle{ACM-Reference-Format}
\bibliography{custom}
\newpage
\section*{Appendix}
\appendix

\section{Dataset}
\label{appendix:Dataset}
The table below provides a summary of the sizes of African speech datasets, presented in ascending order based on the available ASR studies focused on African low-resource languages.
\begin{table}[!h]
\centering
\small
\resizebox{\textwidth}{!}{

\begin{tabular}{p{1cm}p{1.5cm}p{1.5cm}p{3cm}p{1.3cm}}
\hline
\textbf{Authors} & \textbf{Language} & \textbf{Hours (h)} & \textbf{Utterances} & \textbf{Speaker}\\
\hline
\cite{nzeyimana2023kinspeak} & kin & 22,000* & — & — \\
\cite{alabi2024afrihubert} & 43 & 6,551 & — & — \\
\cite{ritchie2022large} & kin, swa & 4,221 & — & — \\
\cite{emezue2025naijavoices} & ibo,hau,yor & 1,838 & 645k & 5,455 \\
\cite{ejigu2024large} & amh & 1,732.5 & — & — \\
\cite{olatunji2023afrispeech} & Afri-eng & 200.9 & 67,577 & 2,463 \\
\cite{afonja2024performant}& Afri-en & 200 & — & 2,300 \\
\cite{doumbouya2021using} & 10 & 10,225 & — & 49 (VA) \\
\cite{mukiibi2022makerere} & lin & 155 & — & — \\
\cite{asfaw} & amh & 131 & $\sim$1200 & — \\
\cite{demisse2024amharic} & amh & 110 & 29,221 & 214 \\
\cite{peterson2022openasr21}& 15 & 25/lang & — \\
\cite{sikasote2022bembaspeech} & bem & 24.5 & 14,438 & 17 \\
\cite{tachbelie2020analysis} & amh & 20.8 & 10,850 & 100 \\
\cite{abubakar2024development} & hau & 22 & 5,362 & — \\
\cite{ibrahim2022development} & dag & 9.6 & 11,207 & 22 \\
\cite{jacobs2023towards} & swa, wol & 16.5 & swa 8941, wol 2000 & — \\
\cite{sanni2025afrispeech} & Afri-en & 7 & — & 50 \\
\cite{babatunde2023automatic}  & pcm  & 6.8 & 4,278 & — \\
\cite{kimanuka2024speech} & lin & 4.3 & 2,849 & 32 \\
\cite{ayall2024amharic}  & amh & $\sim$2–3 & 12,000 & 120 \\
\cite{stan2022lightweight} & ban & $\sim$0.1 & 104 & 26 \\
\hline
\end{tabular}
}
\caption{Overview of ASR Datasets for African Languages, * is indicates unlabeld data, Afri-en is African-accented English}
\label{tab:dataset}
\end{table}
\newpage
\subsection{Categorization of Dataset Sources by Domain, Type, and Distribution Platform}
\label{appendix:DatasetSources}
\begin{table}[!h]
\centering
\small
\resizebox{\textwidth}{!}{
\begin{tabular}{p{0.2cm}p{2.5cm}p{2.5cm}p{11cm}}
\hline
\textbf{\#} & \textbf{Category} & \textbf{Category Type} & \textbf{Sources} \\
\hline
1 & Research Programs & Domain/Institution & Collected via LIG-Aikuma app during AMMI program, Masakhane “AfriNames” project, NIST/IARPA MATERIAL program, AIMSammi, Masakhane, Univ. of Pretoria, UN Global Pulse project (Uganda), VoxLingua107 + Nybaiboly.net Bible readings, OpenSLR (publicly available), ASR low-resource toolkit experiments, ALFFA Project, KenCorpus project \\
2 & Tech/Community & Domain/Institution & Hugging Face (Emezue et al.), Community (YouTube + volunteers), Spontaneous speech, Crowd-sourced, YouTube, Crowdsourced read speech, Google Research + British Library, Kinshasa (DRC), collected with smartphones, Recorded from 120 volunteers, In-house collection + Mozilla Common Voice \\
3 & Academic Institutions & Domain/Institution & Read speech, Addis Ababa Univ., University of Bremen, Addis Ababa University, CSL, Read speech, Studio read speech, University of Bremen \& Addis Ababa University, Lagos Univ. + North-West Univ., SADiLaR, Jimma University (Command \& Control), Stellenbosch Univ. (South Africa), Univ. Dar es Salaam + Beijing Institute of Technology, High-quality read speech, Stellenbosch Univ. \& UN partners, Read speech recordings, Addis Ababa Institute of Technology (AAiT), University of Zambia \& George Mason University \\
4 & Broadcast \& Media & Domain/Institution & Radio, Broadcast station archives + YouTube, Broadcast news, interviews, public web media, Broadcast/news + read speech \\
5 & Faith Sources & Domain/Institution & Religious read speech, Derived from Open.Bible project (Biblica), Ika New Testament Bible recordings, Jehovah’s Witnesses website (JW.ORG) \\
6 & Project-specific & Content Type & Collected via LIG-Aikuma app during AMMI program, Hugging Face (Emezue et al.), Masakhane “AfriNames” project, NIST/IARPA MATERIAL program, AIMSammi, Masakhane, Univ. of Pretoria, Google Research + British Library, UN Global Pulse project (Uganda), OpenSLR (publicly available), ASR low-resource toolkit experiments, ALFFA Project, KenCorpus project \\
7 & Crowdsourced & Content Type & Community (YouTube + volunteers), Crowd-sourced, YouTube, Crowdsourced \\
8 & Read Speech & Content Type & Read speech, Addis Ababa Univ., Religious read speech, University of Bremen, Addis Ababa University, CSL, Derived from Open.Bible project (Biblica), Studio read speech, University of Bremen \& Addis Ababa University, Ika New Testament Bible recordings, Lagos Univ. + North-West Univ., SADiLaR, Stellenbosch Univ. (South Africa), Univ. Dar es Salaam + Beijing Institute of Technology, High-quality read speech, Stellenbosch Univ. \& UN partners, VoxLingua107 + Nybaiboly.net Bible readings, Read speech recordings, Addis Ababa Institute of Technology (AAiT), University of Zambia \& George Mason University, In-house collection + Mozilla Common Voice, Jehovah’s Witnesses website (JW.ORG) \\
9 & Spontaneous & Content Type & Spontaneous speech, Jimma University (Command \& Control), Kinshasa (DRC), collected with smartphones, Recorded from 120 volunteers \\
10 & Broadcast \& Media & Content Type & Radio, Broadcast station archives + YouTube, Broadcast news, interviews, public web media, Broadcast/news + read speech \\
11 & Project Archive & Distribution Platform & Collected via LIG-Aikuma app during AMMI program, Masakhane “AfriNames” project, Religious read speech, NIST/IARPA MATERIAL program, Derived from Open.Bible project (Biblica), AIMSammi, Masakhane, Univ. of Pretoria, Ika New Testament Bible recordings, UN Global Pulse project (Uganda), ASR low-resource toolkit experiments, ALFFA Project, KenCorpus project, Jehovah’s Witnesses website (JW.ORG) \\
12 & Public Repository & Distribution Platform & Hugging Face (Emezue et al.), Google Research + British Library, VoxLingua107 + Nybaiboly.net Bible readings, OpenSLR (publicly available), In-house collection + Mozilla Common Voice \\
13 & Community-driven & Distribution Platform & Community (YouTube + volunteers), Spontaneous speech, Radio, Broadcast station archives + YouTube, Broadcast news, interviews, public web media, Crowd-sourced, YouTube, Crowdsourced read speech, Kinshasa (DRC), collected with smartphones, Recorded from 120 volunteers, Broadcast/news + read speech \\
14 & Institutional Archive & Distribution Platform & Read speech, Addis Ababa Univ., University of Bremen, Addis Ababa University, CSL, Studio read speech, University of Bremen \& Addis Ababa University, Lagos Univ. + North-West Univ., SADiLaR, Jimma University (Command \& Control), Stellenbosch Univ. (South Africa), Univ. Dar es Salaam + Beijing Institute of Technology, High-quality read speech, Stellenbosch Univ. \& UN partners, Read speech recordings, Addis Ababa Institute of Technology (AAiT), University of Zambia \& George Mason University \\
\hline
\end{tabular}
}
\caption{Categorization of Dataset Sources by Domain, Content Type, and Distribution Platform}
\label{tab:dataset_sources}
\end{table}

\newpage
\subsection{Linguistic Coverage in African Speech Datasets}
\label{appendix:Linguistic}
The table below presents a list of languages and dialects represented in speech datasets for African low-resource languages.

\begin{table}[!h]
\centering
\small
\begin{tabular}{p{3cm}p{5cm}p{0.5cm}p{3cm}}
\hline
\textbf{Language} & \textbf{Language Family} & \textbf{\# } & \textbf{Examples of Datasets} \\
\hline
Amharic (\texttt{amh}) & Afroasiatic (Semitic) & 15 & \cite{tachbelie2020dnn, ejigu2024large, asfaw, demisse2024amharic, abdou2024multilingual}\\
Yorùbá (\texttt{yor})  & Niger-Congo (Volta-Niger) & 9 & \cite{emezue2025naijavoices, gutkin2020developing, ogunremi2023r, dossou2021okwugbe, meyer2022bibletts}\\
Hausa (\texttt{hau})  & Afroasiatic (Chadic) & 7 & \cite{emezue2025naijavoices, abubakar2024development, ibrahim2022development, meyer2022bibletts}\\
Swahili (\texttt{swa})  & Niger-Congo (Bantu) & 6 & \cite{jacobs2023towards, peterson2022openasr21, kivaisi2023swahili}\\
Igbo (\texttt{ibo})  & Niger-Congo (Volta-Niger) & 4 & \cite{emezue2025naijavoices, dossou2021okwugbe} \\
Tigrigna (\texttt{tir})  & Afroasiatic (Semitic) & 4 & \cite{abate2021endtoend, abate2020large, abate2020deep, tachbelie2020analysis}\\
Oromo (\texttt{orm})  & Afroasiatic (Cushitic) & 4 & \cite{tachbelie2020dnn, abate2020deep} \\
Wolof (\texttt{wol})  & Niger-Congo (Senegambian) & 4 & \cite{jacobs2023towards, mohamud2021fast} \\
Kinyarwanda (\texttt{kin})  & Niger-Congo (Bantu) & 3 & \cite{ritchie2022large, nzeyimana2023kinspeak} \\
Lingala (\texttt{lin})  & Niger-Congo (Bantu) & 3 & \cite{kimanuka2024speech, kabongo2022listra}\\
Dagbani (\texttt{dag}) & Niger-Congo (Gur) & 1 & \cite{ibrahim2023breaking} \\
Fon (\texttt{fon}) & Niger-Congo (Gbe) & 1 & \cite{dossou2021okwugbe}\\
Tshiluba (\texttt{lua}), Kikongo(\texttt{kon}) & Niger-Congo (Bantu) & 1 & \cite{kimanuka2024speech}\\
Bambara (\texttt{bam}) & Niger-Congo (Mande) & 2 & \cite{stan2022lightweight, westhuizen2021multilingual}\\
Fulfulde (\texttt{ful}) & Niger-Congo (Atlantic-Congo) & 1 & \cite{westhuizen2021multilingual}\\
Somali (\texttt{som}) & Afroasiatic (Cushitic) & 2 & \cite{mohamud2021fast}\\
Twi (\texttt{twi}) & Niger-Congo (Kwa) & 1 & \cite{meyer2022bibletts}\\
Luganda (\texttt{lug}) & Niger-Congo (Bantu) & 2 & \cite{mukiibi2022makerere, meyer2022bibletts}\\
Chichewa (\texttt{nya}), Ewe (\texttt{ewe}), Kikuyu (\texttt{kik}) & Niger-Congo (Bantu/Kwa) & 1 & \cite{meyer2022bibletts}\\
Hadasiyya (\texttt{had}) & Afroasiatic (?) / Isolate & 1 & \cite{shamore2023hadiyyissa}\\
Shona (\texttt{sna}) & Niger-Congo (Bantu) & 1 & \cite{sirora2024shona}\\
\hline
\end{tabular}
\caption{Language Coverage and Dataset Examples for African Speech Resources}
\label{tab:Linguistic}
\end{table}

\newpage
\subsection{Annotation Quality}
\label{appendix:annotation quality}
The table below presents a comprehensive overview of the quality of annotations for African low-resource languages.
\begin{table}[!h]
\centering
\small
\begin{tabular}{p{1cm}p{2.5cm}p{2cm}p{7cm}}
\hline
\textbf{Papers} & \textbf{Anno. Type} & \textbf{Granularity} & \textbf{Description} \\
\cite{tachbelie2020analysis} & Manual orthographic & Sentence, phoneme & Fully aligned with pronunciation lexicons, High-quality, clean lab conditions, 43-phone lexicon \\
\cite{emezue2025naijavoices}. & Double human review & Sentence-level & Provided in train/val/test manifests, Designed for reproducibility and multilingual benchmarks \\
\cite{ejigu2024large, asfaw} & Manual transcription & Word \& sentence-level & CTC alignment (unsupervised), Used for large-scale Amharic model training, no forced aligner \\
\cite{demisse2024amharic} & Verified double entry & Prompt-level & Grapheme and subword alignment, Built for both CER and WER evaluation \\
\cite{abubakar2024development} & Human-validated clips & Clip-level (Common Voice) & Diacritic-aware DER + WER metrics, Focus on tonal accuracy in Hausa \\
\cite{peterson2022openasr21} & NIST reference transcripts & Case-sensitive/insensitive WER & Blind EVAL partitions, Alignments conform to strict evaluation protocols \\
\cite{gutkin2020developing} & Fully diacritized, sentence-aligned & Sentence-level & With Praat TextGrid timestamps, Includes Yoruba tones, CMU Wilderness compatibility \\
\cite{dossou2021okwugbe} & Manual transcripts & Word-level & No forced alignment, Kaldi recipes provided for reproducibility \\
\cite{ritchie2022large} & Mixed (inherited annotations) & Varies by source & \~20\% held out for eval per language, Aggregated corpora, annotations not normalized across all datasets \\
\cite{kivaisi2023swahili} & Manually verified digits & Word-level & No timestamps, MIT-licensed Swahili digits corpus \\
\cite{abate2020deep} & Automatic pronunciation generation & Lexical & Aligned to Kaldi recipes, From 4M-token corpus for Amharic/Tigrigna \\
\cite{biswas2022code} & Manual with code-switching tags & Word-level + switch tags & 100ms segmentation, High linguistic richness, used for multilingual WER evaluations \\
\bottomrule
\end{tabular}
\caption{Overview of annotation quality, type, granularities, and alignment strategies across datasets}
\end{table}
\newpage
\subsection{Dataset Licenses, Splits, and benchmarks}
\label{appendix:licenses}
The tables below detail the licensing of datasets, the availability of fixed techniques, and the benchmarking for low-resource languages in Africa.
\begin{table}[!h]
\centering
\small
\begin{tabular}{p{1cm}p{2cm}p{2cm}p{2cm}p{5cm}}
\hline
\textbf{Papers} & \textbf{License} & \textbf{Split Provided?} & \textbf{Benchmark Protocol?} & \textbf{Description} \\
\hline
\cite{emezue2025naijavoices} & CC BY-NC-SA 4.0 & train/val/test & JSON manifests for benchmarking & Large-scale, speaker-diverse corpus for Hausa, Yoruba, and Igbo \\
\cite{tachbelie2023lexical} & CC BY-NC-SA 3.0 (research only) & train/dev/test & Kaldi + base models provided & Standard benchmark protocol used in Amharic ASR \\
\cite{abubakar2024development} & CC0 (Mozilla Common Voice) & train/val/test & WER + DER evaluated & Fully open and diacritic-aware Hausa dataset \\
\cite{demisse2024amharic} & ALFFA (academic license) & fixed splits & CER and WER benchmarks & Detailed transcripts and subword units for Amharic \\
\cite{asfaw} & No license stated & (internal split only) & Internal test set only & Used CTC optimizer but lacks public protocol \\
\cite{nzeyimana2023kinspeak} & Public domain (YouTube + JW) & dev/test provided & Not benchmarked publicly & Large SSL-friendly set, but no formal benchmark \\
\cite{biswas2022code} & Research use only & balanced + full & multilingual WER evaluation & Code-switching labels + multiple splits \\
\cite{afonja2024performant} & CC BY-NC-SA 4.0 & train/dev/test & used for NER + ASR & Multitask African-accented English corpus \\
\cite{dossou2021okwugbe} & CC BY-NC-SA 4.0 & dev/test & Kaldi recipes on GitHub & Fon + Igbo corpus with reproducible splits \\
\cite{peterson2022openasr21} & NIST OpenSAT challenge license & fixed BUILD/DEV/EVAL & blind EVAL protocol & Competitive benchmark design for low-resource languages \\
\cite{ritchie2022large} & Mixed open licenses & \~20\% held out & each dataset retains original setup & Aggregated corpora, non-unified benchmark \\
\cite{gutkin2020developing} & CC BY 4.0 & train/dev/test & compatible with CMU Wilderness & Yoruba speech with full diacritics + timestamped alignment \\
\cite{kimanuka2024speech} & CC BY & train/dev/test & benchmark not stated & Lingala; simple, ready-to-use setup \\
\cite{ibrahim2023breaking} & CC0 (Wikimedia Commons) & 70/20/10 & research only & Dagbani speech from smartphones \\
\cite{westhuizen2021multilingual} & Research-only via AAU & standard split & reused in other Kaldi models & Bambara and Fulfulde from radio talk shows \\
\hline
\end{tabular}
\caption{Overview of dataset licenses, splits, benchmarks, and notes across various African speech corpora}
\end{table}

\newpage

\newpage
\section{Models Architecture}
\subsection{Training Techniques}
\label{appendix:Training_Tech}
The table below showcases various techniques employed by researchers in ASR for African low-resource languages.
\begin{table*}[!h]
\centering
\small
\begin{tabular}{p{1cm}p{6cm}p{5.5cm}}
\hline
\textbf{Authors} & \textbf{Training Technique\& Description} & \textbf{Remarks} \\ \hline

\cite{abubakar2024development,alabi2024afrihubert,ramanantsoa2023voxmg,olatunji2023afrinames} & \textit{SSL:} Pretraining on unlabeled data , followed by supervised finetuning. &  Highly effective in data-scarce settings. However, pretraining requires large pretraining datasets. \\ 

\cite{nzenwata2024ika,ogunremi2023r,rikhotso2021development,asfaw}& \textit{Supervised Finetuning:} Training on labeled data using standard loss functions, often with pretrained models &  Key technique in most papers, and its success varies based on data quality and size. Overfitting is a risk.\\ 

\cite{sikasote2022bembaspeech,westhuizen2021multilingual,el2023comparative,nzenwata2024ika,padhi2020multilingual}& \textit{TL:} Transferring pretrained models from high-resource to low-resource languages or domains & Allows quick adaptation and careful language selection, which is significantly affects success.\\ 

\cite{abate2021endtoend,abate2020deep,alabi2024afrihubert,fantaye2020investigation,emiru2021improving,padhi2020multilingual}& \textit{Multilingual Training:} Training together on different related/unrelated lang. to improve generalization and reduce OOVs &  Effective for similar lang. and can cause confusion if phoneme or orthographic overlap is not handled. \\ 

 \cite{nzeyimana2023kinspeak}&Curriculum Learning \& Staging training from clean to noisy or from easier to harder tasks & Promotes noise robustness, especially useful when mixing clean studio speech and noisy field data. Adds training complexity but improves convergence. \\ 

\cite{nzeyimana2023kinspeak,mohamed2023multilingual,jacobs2023towards}& \textit{Semi-SL:} Using unlabeled data with multi-generation bootstrapping &  Improve performance in small data settings. Combined with SSL for high performance. \\ 

\cite{adnew2024semantically,afonja2024performant,peterson2022openasr21,emiru2021improving,padhi2020multilingual,nzeyimana2023kinspeak,olatunji2023afrinames,jacobs2023towards}& \textit{Data Augmentation:} Artificially expanding training data using noise addition & Essential for robustness and generalization, especially for accent or dialect diversity. Widely used across papers. \\ 

\cite{emiru2021improving,abate2020deep,teshite2023afan,nzeyimana2023kinspeak,tachbelie2020analysis}& \textit{Tokenization Strategies:} Varying modeling units to better reflect language structure, especially for agglutinative or tonal languages &  Morpheme and phoneme help reduce OOVs and linguistic structure, trade-off is increased complexity in lexicon and G2P mapping. \\ 

\cite{adnew2024semantically,emiru2021improving,padhi2020multilingual,jimerson2023unhelpful,kabongo2022listra} &  \textit{Loss Function Variants:} Using relevant loss functions such as CTC, cross-entropy, or LF-MMI & CTC remains dominant due to alignment flexibility and Seq2Seq adds contextual modeling. LF-MMI is prominent in Kaldi pipelines. \\ 
\hline
\end{tabular}
\caption{Training techniques}
\end{table*}

\newpage

\newpage
\subsection{Evaluation Metrics}
The table below provides information on evaluation metrics used to assess the ASR for African low-resource languages.
\label{appendix:asr_metrics}
\begin{table}[htbp]
\centering
\begin{threeparttable}
\begin{tabular}{p{2cm}p{2cm}p{8cm}}
\toprule
\textbf{Authors} & \textbf{Eval. Metric} & \textbf{Description \& Remarks} \\
\midrule
\cite{abubakar2024development, adnew2024semantically, asfaw, emezue2025naijavoices, ramanantsoa2023voxmg, emiru2021improving, babatunde2023automatic} 
& WER 
&  Widely used for ASR accuracy benchmarking. \& Dominant metric across all papers. Enables comparisons but may not reflect semantic or named-entity accuracy. \\
\cite{adnew2024semantically, tachbelie2023lexical, nzenwata2024ika}  
& CER 
& Measures errors at the character level and is useful in languages with complex morphology or limited word segmentation. \& Important for morphologically rich languages. \\
\cite{abubakar2024development, shamore2023hadiyyissa}  
& DER 
& Measures diacritic restoration accuracy and is critical for tonal languages \& Essential in diacritic-sensitive languages. Missing from most general ASR evaluations. \\
\cite{ayall2024amharic, el2023comparative, teshite2023afan, stan2022lightweight, emezue2023afrodigits} 
& Accuracy 
& Employed in limited vocabulary. Suitable for recognizing fixed words but doesn't represent real world performance. \\
\cite{afonja2024performant} 
& MWER 
& focused on NER in medical speech \& Highlights inconsistency between general ASR and domain-specific accuracy \\
\cite{adnew2024semantically, olatunji2023afrinames, kabongo2022listra} 
& Semantic Correctness
& Measures output fluency \& Not standardized. \\
\cite{kabongo2022listra}
& BLEU Score 
& Uses n\mbox{-}gram to compare translation output to a reference 
\& Applicable only to ASR-to-MT pipelines, and the score depends on the reference quality and fluency. \\
\cite{asfaw, westhuizen2021multilingual, alhumud2024improving, olatunji2023afrinames}
& Cross-dialect 
& Evaluates systems across dialects 
\& Increasingly important but few researches include it. \\
\cite{teshite2023afan, stan2022lightweight, jacobs2023towards} 
& Live vs.\ Non-live Testing 
& Evaluates models in real-time  
\& Useful for deployment readiness and reveals speaker variation. \\
\bottomrule
\end{tabular}
\caption{Evaluation Metrics}
\end{threeparttable}
\end{table}
\newpage
\subsection{ASR Strategies and focus area}
The table below offers insights into various strategies used by researchers working on ASR for African low-resource languages. 
\label{appendix:asr_strategies}
\begin{table}[htbp]
\centering
\begin{threeparttable}
\begin{tabular}{p{1.5cm}p{3cm}p{9cm}}

\toprule
\textbf{Authors} & \textbf{Technique} & \textbf{Description \& Remarks} \\
\midrule
\cite{asfaw,westhuizen2021multilingual, alhumud2024improving, oladipo2020accent, olatunji2023afrinames} 
& Dialect and Accent control 
& Adapts to minimize bias and WER. It shows significant improvement \\
\addlinespace
\cite{tachbelie2020dnn, woldemariam2020transfer, westhuizen2021multilingual, fantaye2020investigation, abate2021endtoend}  
& Code switching 
& Merges close lang. using TL to improve low-resource lang. Powerful when lang. are close. \\
\addlinespace
\cite{alabi2024afrihubert, padhi2020multilingual, emiru2021improving} 
& Multilingual Pretraining  
& Trains on different lang. for better representations. This increases robustness to OOVs and domain shifts. \\
\addlinespace
\cite{nzeyimana2023kinspeak, asfaw, mukiibi2022makerere, jacobs2023towards, peterson2022openasr21, alhumud2024improving}
& Noise Robustness  
& Uses in recordings to improve robustness. Noise augmentation reduces domain inconsistency, but it may increase training complexity. \\
\addlinespace
\cite{abate2020large, emiru2021improving, teshite2023afan, tachbelie2023lexical} 
& Morpheme 
& Reduces OOV. Improves adaptability for high-morphology languages. \\
\addlinespace
\cite{alhumud2024improving, olatunji2023afrinames, oladipo2020accent} 
& Accent-Aware TL / Finetuning 
& Finetunes models improve entity recall and accuracy. \\
\addlinespace
\cite{emezue2025naijavoices, ogunremi2023r, afonja2024performant, meyer2022bibletts} 
& Domain Diversity 
& Trains on different speakers, genders, and domains. Boosts generalization and reduces overfitting. \\
\addlinespace
\cite{teshite2023afan, stan2022lightweight, jacobs2023towards, mukiibi2022makerere} 
& Field Deployment Testing 
& Evaluates systems in reality. Validates robustness. \\
\addlinespace
\cite{afonja2024performant, nzeyimana2023kinspeak, alhumud2024improving} 
& Cross-Domain Adaptation 
& Adapts models trained on one domain to another. Domain shifts are challenging and retrieval-based voice conversion served as solution technique \\
\bottomrule
\end{tabular}
\caption{Strategies and Focus Areas for Robust African ASR}
\end{threeparttable}
\end{table}

\end{document}